**Evaluation and selection of Medical Tourism sites: A rough AHP based MABAC approach**

Jagannath Roy, Kajal Chatterjee, Abhirup Bandhopadhyay, Samarjit Kar
Department of Mathematics, National Institute of Technology, Durgapur, India
jaga.math23@gmail.com, chatterjeekajal7@gmail.com, abhirupnit@gmail.com,
dr.samarjitkar@gmail.com

## Abstract

This paper presents a novel multiple **c**riteria **d**ecision **m**aking (MCDM) methodology for assessing and prioritizing medical tourism destinations in uncertain environment. A systematic evaluation and assessment method is proposed by incorporating Analytic Hierarchy Process (AHP) and Multi-Attribute Border Approximation area Comparison (MABAC) methods in rough number. Rough number effectively aggregate individual judgments of decision makers and express their true perception to handle vagueness without any prior information. Rough AHP analyzes the relative importance of criteria based on their preferences given by experts while Rough MABAC evaluates the alternative sites based on the criteria weights. Finally, an application in prioritizing different cities in India for medical tourism service is proposed to demonstrate the method. The validity of the obtained ranking for the given decision making problem is established by testing criteria proposed by Wang and Triantaphyllou (2008) along with comparison with rough TOPSIS and rough VIKOR.

## 1. Introduction

Globalization of trade in the health service has given a new form in tourism sector-‘*Medical tourism’.* High costs of treatment, long waiting time, affordability of airfares to overseas destinations and favorable exchange rate change are crucial factors related to the fast growth of Medical Tourism (Connell, 2006). Rapid development of medical infrastructure with international standards and certification, easy availability of skilled manpower bring South Asian countries like Thailand, Malaysia, and India at the forefront in this area. With current annual growth of 13.0 percent, the Indian health care sector contributes about $ 23 billion (*nearly 4 percent of GDP*) to the Indian economy, with ‘*foreign exchange earning around $1.8 billion*’ (Chakraborty, 2006). Although research studies are abundant focusing on social impacts of Medical Tourism, there is no proper methodology for customers, both foreign and domestic, to assess the medical tourist destination in any country. The problem can be solved by taking the interest of stakeholder’s in assessing the weights of a multiple criteria set, *namely medical infrastructure, logistics service providers,* *government policy along with city demography*. Therefore, assessment of desirable medical destination selection and evaluation problem can be considered decision making problem with multiple attributes varying from consumer demands to resource constraints of medical related industry.

In this regard, MCDM has become a very crucial area of management research and decision theory with lots of methods developed, extended and modified in solving problems in the present and past few decades. Some of them are namely, MABAC (Pamucar and Cirovic, 2015), TOPSIS (*Technique for Order Preference by Similarity to Ideal Solution*) (Hwang and Yoon, 1981), AHP (Saaty, 1977, 1990), ANP (*Analytic Network Process*) (Saaty, 1996), DEMATEL (*Decision Making Trial and Evaluation Laboratory*) (Gabus and Fontela, 1972), PROMETHEE (Preference *Ranking Organization METHods for Enrichment Evaluations*) (Brans et al.1984, 1985), ELECTRE (*Elimination Et Choice Translating REality*) (Roy, 1968), DRSA (*Dominance based Rough Set Approach*) (Greco et al. 2001), VIKOR (*VIsekriterijumska optimizacija KOmpromisno Resenje*) (Opricovic, 1998), DEA (*Data Envelopment Analysis*) (Charnes et al. 1978). Each of them has some advantages and disadvantages, yet they provide satisfactory optimal (*compromised*) solutions to the given problem.

Earlier researchers dealt with precise and certain information based MADM methods. As per Khoo et al. (1999) decision support is based on human knowledge about a specific part of a real or abstract world problem. Since human decisions are uncertain and vague, decision makers give their preferences in linguistic terms instead of deterministic value. Many theories and techniques are developed to analyse human subjective judgment based on imprecise data. Some of them are fuzzy sets theory (Zadeh, 1965), Dempster–Shafer theory of evidence (Shafer, 1976), grey theory (Julong, 1989), rough set theory (Pawlak, 1982). Rough set theoretic approach is one of the best choices to solve such uncertain MCDM problem (Pawlak, 1982). Rough number, derived from the basic notion of approximations in rough set theory, mainly aggregate expert's judgments in limited data and uncertain information. Greco et al (2001) introduced DRSA (*Dominance based Rough Set Approach*) to overcome the barriers but it also has some disadvantages since dominance relation is very weak relation which leaves many objects *(alternatives)* uncompared. Zhai et al. (2008) defined '*rough numbers and rough boundary interval'* through the use of upper and lower approximations, extended form of basic rough set theory. Due to difficulty in determining appropriate fuzzy membership function and boundary intervals, rough number is used to measure the vagueness in many decision making problems (Zhu et al.2015).

In this paper, we extend MABAC method for rough numbers and hybridized with rough number based Analytic Hierarchy Process (R-AHP) for MCDM problem, where degree of imprecision is not pre-assumed as in fuzzy theory or probability theory. The proposed hybrid method produces all results just from a given data set and no auxiliary information is needed. In this paper, a hybrid AHP-MABAC method (see fig.2) dealing with rough numbers is developed to assist decision makers for evaluation and assessment of the optimal alternative(s) for an MCDM problem.

The paper is structured as follows. Section 2 introduces the basic concepts on the rough numbers, AHP and MABAC methods. Section 3 presents the proposed hybrid AHP-MABAC method based on rough numbers. The implementation of the proposed hybrid method for evaluating the medical tourism sites in India is provided in section 4. Comparative analysis and validity testing of proposed method is done in section 5. Finally, section 6 summarizes the paper.

## 2. Preliminaries

### *2.1 Rough numbers*

Rough number (Zhai et al. 2008) based on rough sets was developed in determining the boundary interval to handle subjective judgment of decision makers. It was further integrated with interval arithmetic operations to analyse vague information (Zhai et al., 2009). A rough number with lower, upper and boundary interval respectively, does not require any subjective adjustment to analyse data. Unifying rough number in concept evaluation structure, decision makers will give rational decisions in subjective situation. Assume that $\Lambda$ be the universe of all the objects, $X$ an arbitrary object of $\Lambda$ and $E$ a set of $t$ class that covers all objects in $\Lambda$ i.e.

$$R = \{\mathcal{G}_1, \mathcal{G}_2, \ldots, \mathcal{G}_t\} \text{ then } \mathcal{G}_t \in E \quad \forall\, X \in \Lambda,\ 1 \le q \le t \text{ provided } \mathcal{G}_1 < \mathcal{G}_2 <, \ldots, < \mathcal{G}_t$$

The lower approximation($\underline{Apr}(\mathcal{G}_q)$), upper approximation ($\overline{Apr}(\mathcal{G}_q)$) and boundary region ($Bnd(\mathcal{G}_q)$) of class $\mathcal{G}_q$ are defined as:

$$\underline{Apr}(\mathcal{G}_q) = \cup\{X \in \Lambda : E(X) \le \mathcal{G}_q\} \tag{1}$$

$$\overline{Apr}(\mathcal{G}_q) = \cup\{X \in \Lambda : E(X) \ge \mathcal{G}_q\} \tag{2}$$

$$Bnd(\mathcal{G}_q) = \left\{X \in \Lambda : E(X) < \mathcal{G}_q\right\} \cup \{X \in \Lambda : E(X) > \mathcal{G}_q\} \tag{3}$$

Then class $\mathcal{G}_q$ denoted by $RN(\mathcal{G}_q)$ with corresponding lower limit ($\underline{Lim}(\mathcal{G}_q)$) and upper limit ($\overline{Lim}(\mathcal{G}_q)$) as follows:

$$RN(\mathcal{G}_q) = \left[\underline{Lim}(\mathcal{G}_q), \overline{Lim}(\mathcal{G}_q)\right] \tag{4}$$

where

$$\underline{Lim}(\mathcal{G}_q) = \frac{1}{M_L}\sum E(X) \mid X \in \underline{Apr}(\mathcal{G}_q) \tag{5}$$

$$\overline{Lim}(\mathcal{G}_q) = \frac{1}{M_U}\sum E(X) \mid X \in \overline{Apr}(\mathcal{G}_q) \tag{6}$$

$M_L$, $M_U$ belongs to $\underline{Apr}(\mathcal{G}_q)$ and $\overline{Apr}(\mathcal{G}_q)$ respectively. The lower $\underline{Lim}(\mathcal{G}_q)$ and the upper limit $\overline{Lim}(\mathcal{G}_q)$ denotes the mean value of elements included in its corresponding lower and upper approximation, respectively, with their difference denoted as rough boundary interval $IRBnd(\mathcal{G}_q)$, in (7).

$$IRBnd(\mathcal{G}_q) = \overline{Lim}(\mathcal{G}_q) - \underline{Lim}(\mathcal{G}_q) \tag{7}$$

### *2.2 Ranking rule for rough numbers*

For any two rough numbers, $RN(\alpha) = \lfloor \underline{Lim}(\alpha), \overline{Lim}(\alpha) \rfloor$ and $RN(\beta) = \lfloor \underline{Lim}(\beta), \overline{Lim}(\beta) \rfloor$, the ranking rule of interval numbers is defined as per Zhai et al. (2008).

(1) The ranking order can be easily found out, if rough boundary interval of rough numbers are not strictly bounded; giving option (a) and (b).

(a) If $\overline{Lim}(\alpha) > \overline{Lim}(\beta)$ and $\underline{Lim}(\alpha) \geq \underline{Lim}(\beta)$, or $\overline{Lim}(\alpha) \geq \overline{Lim}(\beta)$ and $\underline{Lim}(\alpha) < \underline{Lim}(\beta)$ then $RN(\alpha) > RN(\beta)$ (Fig. 1(a)).

(b) If $\overline{Lim}(\alpha) = \overline{Lim}(\beta)$ and $\underline{Lim}(\alpha) = \underline{Lim}(\beta)$, then $RN(\alpha) = RN(\beta)$ (Fig. 1(b)).

(2)The ranking become tedious, if rough boundary interval of $RN(\alpha) = \lfloor \underline{Lim}(\alpha), \overline{Lim}(\alpha) \rfloor$ and $RN(\beta) = \lfloor \underline{Lim}(\beta), \overline{Lim}(\beta) \rfloor$ are strictly bounded thus opening various cases. Here we consider, $M(\alpha)\ and\ M(\beta)$ middle values of $RN(\alpha)$ and $RN(\beta)$ respectively.

If $\overline{Lim}(\alpha) > \overline{Lim}(\beta)$ and $\underline{Lim}(\alpha) < \underline{Lim}(\beta)$, three cases arise depending on values of $M(\alpha)$and $M(\beta)$:

(a) If $M(\alpha) < M(\beta)$, then $RN(\alpha) < RN(\beta)$ (Fig. 1(c)).

(b) If $M(\alpha) = M(\beta)$, then $RN(\alpha) < RN(\beta)$ (Fig. 1(d)).

(c) If $M(\alpha) > M(\beta)$, then $RN(\alpha) > RN(\beta)$ (Fig. 1(e)).

### *2.3 Interval arithmetic of rough numbers*

Although possessing different characteristics, both rough numbers and fuzzy numbers share similar mathematical implications. Both of them can be used to describe vague information, and the degree of vagueness is measured by the range of boundary mathematics, where the concept of 'intervals' (*also known as 'bounds'*) consisting of lower and upper limits, describe vague data. Interval mathematics also provides a set of arithmetic operations for interval computation, including the interval arithmetic. Due to its similarity with interval number, the arithmetic rules of interval numbers can also be applied in rough numbers (Zhai et al, 2009).

The interval arithmetic operation for two rough numbers, $RN(\alpha) = \left[\underline{Lim}(\alpha), \overline{Lim}(\alpha)\right]$ and $RN(\beta) = \left[\underline{Lim}(\beta), \overline{Lim}(\beta)\right]$, is carried out as per (8)-(12).

1. Addition (+) of two rough numbers $RN(\alpha)$ and $RN(\beta)$

$$RN(\alpha) + RN(\beta) = \left[\underline{Lim}(\alpha) + \underline{Lim}(\beta), \overline{Lim}(\alpha) + \overline{Lim}(\beta)\right] \tag{8}$$

2. Subtraction (-) of two rough numbers $RN(\alpha)$ and $RN(\beta)$

$$RN(\alpha)(-)RN(\beta) = \left[\underline{Lim}(\alpha) - \overline{Lim}(\beta), \overline{Lim}(\alpha) - \underline{Lim}(\beta)\right] \tag{9}$$

3. Multiplication (×) of two rough numbers $RN(\alpha)$ and $RN(\beta)$

$$RN(\alpha) \times RN(\beta) = \left[\underline{Lim}(\alpha) \times \underline{Lim}(\beta), \overline{Lim}(\alpha) \times \overline{Lim}(\beta)\right] \tag{10}$$

4. Division (÷) of two rough numbers $RN(a)$ and $RN(b)$

$$RN(\alpha) \div RN(\beta) = \left[\underline{Lim}(\alpha) \div \overline{Lim}(\beta), \overline{Lim}(\alpha) \div \underline{Lim}(\beta)\right] \tag{11}$$

5. Scalar multiplication of rough number $RN(\alpha)$, where $\mu$ is a nonzero constant

$$\mu \ RN(\alpha) \quad = \left[\mu \, \underline{Lim}(\alpha), \ \mu \, \overline{Lim}(\alpha)\right] \tag{12}$$

## 3 Proposed Methodology

This study aims at combining the AHP and MABAC method with rough number to evaluate alternatives under imprecise environment under group decision-making problem. First, rough AHP is developed for weight determination of evaluation criteria. Second, modified MABAC model based on rough number is proposed to evaluate the alternatives. The details are discussed below.

### *2.2 Rough AHP approach for criteria weighting*

Rough AHP (Song et al. 2013) measure consistency of preferences while managing and manipulating decision-making involving subjective judgments. This proposed paper combines rough AHP where rough number handles subjectivity and AHP handles hierarchy evaluation. The detailed procedure is as follows (Zhu et al. 2015)

***Step 1:*** Form a hierarchy for evaluation criteria. A committee of $k$ experts is formed to select the criteria and obtain the prospective alternatives for the decision making problem.

***Step 2:*** Develop a group of pair-wise comparison matrix. The expert team are invited to make pair-wise comparison of criteria to obtain priority weights of data matrix. The $e^{th}$ expert pairwise comparison matrix $B_e$ expressed as:

$$B_e = \begin{bmatrix} 1 & b_{12}^e & \cdots & b_{1n}^e \\ b_{21}^e & 1 & \cdots & b_{2n}^e \\ \vdots & \vdots & \ddots & \vdots \\ b_{n1}^e & b_{n2}^e & \cdots & 1 \end{bmatrix}_{n\times n}, \quad 1 \le i,\ j \le n,\ 1 \le e \le k \tag{13}$$

Thus, $B_1, B_2, \dots, B_k$ are the matrices provided by $k$ experts for $i^{th}$ criterion compared with $j^{th}$ criterion. Calculate the maximum eigenvalue $\lambda_{max}^e$ of $B_e$, then compute the consistency index $(CI)$ as in (14).

$$\left. \begin{aligned} CI &= \frac{\lambda_{\max}^e - n}{n-1} \\ CR^* &= \frac{CI}{RI(n)} \end{aligned} \right\} \tag{14}$$

Determine the random consistency index $(RI)$ (Table 1), and then compute the consistency ratio $CR^*$. If $CR < 0.1$, eqn. (13) is acceptable, otherwise, experts' judgments be properly adjusted. Finally, by aggregating the expert opinions, the integrated comparison matrix $B$ is constructed as:

$$B = \begin{bmatrix} 1 & b_{12} & \cdots & b_{1n} \\ b_{21} & 1 & \cdots & b_{2n} \\ \vdots & \vdots & \ddots & \vdots \\ b_{n1} & b_{n2} & \cdots & 1 \end{bmatrix}_{n\times n} \tag{15}$$

where $b_{ij} = \left\{ b_{ij}^1, b_{ij}^2, \dots, b_{ij}^k \right\}$ , as the sequence of relative importance of criterion $i$ on $j$.

***Step 3:*** Using Eqs. (1)- (6), translate the element $b_{ij}$ in group decision matrix $B$ into $RN\left(b_{ij}^e\right)$ of $b_{ij}$ as:

$$RN\left(b_{ij}^e\right) = \left[b_{ij}^{eL}, b_{ij}^{eU}\right] \tag{16}$$

$b_{ij}^{eL}$ the lower limit and $b_{ij}^{eU}$ the upper limit of rough number $RN\left(b_{ij}^e\right)$ in $e^{th}$ pair-wise comparison matrix respectively.

Then we obtain a rough sequence $RN\left(b_{ij}\right)$ represented in (17) as:

$$RN\left(b_{ij}\right) = \{\left[b_{ij}^{1L}, b_{ij}^{1U}\right], \left[b_{ij}^{2L}, b_{ij}^{2U}\right], \dots, \left[b_{ij}^{kL}, b_{ij}^{kU}\right]\} \tag{17}$$

The average rough interval $\overline{RN(b_{ij})}$ is obtained by using rough arithmetic operations (8-10):

$$\overline{RN(b_{ij})} = \lceil b_{ij}^L, b_{ij}^U \rfloor \tag{18}$$

where

$$\left.\begin{aligned} b_{ij}^L &= \left( b_{ij}^{1L} + b_{ij}^{2L} + \ldots + b_{ij}^{kL} \right) / k \\ b_{ij}^U &= \left( b_{ij}^{1U} + b_{ij}^{2U} + \ldots + b_{ij}^{kU} \right) / k \end{aligned}\right\} \tag{19}$$

Then the rough group decision matrix M is formed as follows:

$$M = \begin{pmatrix} \lceil 1,1 \rfloor & \lceil b_{12}^L, b_{12}^U \rfloor & \ldots & \lceil b_{1n}^L, b_{1n}^U \rfloor \\ \lceil b_{21}^L, b_{21}^U \rfloor & \lceil 1,1 \rfloor & \ldots & \lceil b_{2n}^L, b_{2n}^U \rfloor \\ \ldots & \ldots \quad \ldots & \ldots \quad \ldots & \ldots \\ \lceil b_{n1}^L, b_{n1}^U \rfloor & \lceil b_{n1}^L, b_{n1}^U \rfloor & \ldots & \lceil 1,1 \rfloor \end{pmatrix} \tag{20}$$

***Step 4:*** Calculate the rough based weight $W_i$ of each criterion:

$$W_i = \left( \left[ W_i^L, W_i^U \right] \right) = \left( \left[ \left( \prod_{j=1}^n b_{ij}^L \right)^{1/n}, \left( \prod_{j=1}^n b_{ij}^U \right)^{1/n} \right] \right) \tag{21}$$

$and$ its normalized counterparts ($\omega_i$) by the following equation (22)

$$\omega_i = \left( \left[ \omega_i^L, \omega_i^U \right] \right) = \left( \left[ \frac{W_i^L}{\max(W_i^U)}, \frac{W_i^U}{\max(W_i^U)} \right] \right) \qquad i = 1, 2, \ldots, n. \tag{22}$$

### ***3.2. Applying the Rough number based MABAC to find the best alternative:***

MABAC method (Pamucar and Cirovic, 2015) is a reliable tool for rational decision making due to its simple computation procedure and the *consistency* of solution. The basis of the method is mainly the distance of the criterion function of each alternative from the border approximation area (BAA). It was modified with interval- valued intuitionistic fuzzy considering subjective assessments and objective data for material selection (Xue et al. 2016). In this paper, the authors modify the concept of MABAC to develop a rough number-based framework for evaluating alternatives based on obtained weight coefficients. The steps are detailed below.

*Step 1.* Develop a group decision-making framework by identifying alternatives in respect to each criterion, and translate it into initial rough decision matrix $(X)$ per equations (1)-(11) in the form of $A_i = \left( \left[ x_{i1}^L, x_{i1}^U \right], \left[ x_{i2}^L, x_{i2}^U \right], \ldots, \left[ x_{in}^L, x_{in}^U \right] \right)$. Suppose there are $m$ alternatives represented in form of rough numbers to be evaluated by $n$ criteria,

$$X = \left( \left[ x_{ij}^L, x_{ij}^U \right] \right)_{m \times n}, \quad i = 1, 2, \ldots, m; j = 1, 2, \ldots, \mathrm{n} \tag{23}$$

$\left[x_{ij}^L, x_{ij}^U\right]$ being the value of the $i$-th alternative as per $j$-th criterion, $m$ denotes the number of alternatives and $n$ the number of criteria.

*Step* 2. Applying eqn. (24)-(27), the normalized matrix $N = \left(\left[n_{ij}^L, n_{ij}^U\right]\right)_{m\times n}$ are derived from the initial matrix $(X)$ using the equations (24)-(27).

(a) For Benefit type criteria

$$n_{ij}^L = \frac{x_{ij}^L - x_j^-}{x_j^+ - x_j^-} \quad , \quad n_{ij}^U = \frac{x_{ij}^U - x_j^-}{x_j^+ - x_j^-} \tag{24}$$

(b) For Cost type criteria

$$n_{ij}^L = \frac{x_{ij}^L - x_j^+}{x_j^- - x_j^+}, \qquad n_{ij}^U = \frac{x_{ij}^U - x_j^+}{x_j^- - x_j^+} \tag{25}$$

Where

$$x_j^+ = \begin{cases} \max\limits_{1\le i\le m}\left(x_{ij}^U\right), & \text{for benefit type criteria} \\ \min\limits_{1\le i\le m}\left(x_{ij}^L\right), & \text{for cost type criteria} \end{cases} \qquad j = 1, 2, \dots, n. \tag{26}$$

$$x_j^- = \begin{cases} \min\limits_{1\le i\le m}\left(x_{ij}^L\right), & \text{for benefit type criteria} \\ \max\limits_{1\le i\le m}\left(x_{ij}^U\right), & \text{for cost type criteria} \end{cases} \qquad j = 1, 2, \dots, n. \tag{27}$$

*Step* 3. Based on eqn. (28) calculate the elements of the weighted decision making matrix $(V)$

$$V = \left(\left[v_{ij}^L, v_{ij}^U\right]\right)_{m\times n} \tag{28}$$

$v_{ij}^L = w_L^i \cdot \left(n_{ij}^L + 1\right)$, $v_{ij}^U = w_U^i \cdot \left(n_{ij}^U + 1\right)$. $\left[n_{ij}^L, n_{ij}^U\right]$ are elements of normalized matrix $(N)$, $\left[w_L^i, w_U^i\right]$ the weight coefficients of the criteria.

*Step* 4. Using the geometric aggregation for interval numbers $g_j = \left[g_j^L, g_j^U\right]$, BAA for each criterion is calculated as per equation (29):

$$\begin{cases} G = [g_1, g_2, \dots, g_n] \;\; where\; g_j = \left[g_j^L, g_j^U\right] \\ g_j^L = \left(\prod_{i=1}^m v_{ij}^L\right)^{1/m} \\ g_j^U = \left(\prod_{i=1}^m v_{ij}^U\right)^{1/m} \end{cases} \tag{29}$$

where $\left[v_{ij}^L, v_{ij}^U\right]$ are the elements of (28).

*Step 5.* The Euclidean distance operator (Hennig et al., 2015) for interval numbers is used here for rough numbers to measure the distances of the alternatives from BAA for getting the distance matrix $Q$ as:

$$Q = \left(q_{ij}\right)_{m\times n} = \left(\left[q_{ij}^L, q_{ij}^U\right]\right)_{m\times n}, \text{ where} \tag{30a}$$

$$\begin{cases} q_{ij} = \begin{cases} d_E(v_{ij}, g_j), & if\ RN(v_{ij}) > RN(g_j) \\ -d_E(v_{ij}, g_j), & if\ RN(v_{ij}) < RN(g_j) \end{cases} \text{for benefit type criteria (See Section 2.2)} \\ q_{ij} = \begin{cases} -d_E(v_{ij}, g_j), & if\ RN(v_{ij}) > RN(g_j) \\ d_E(v_{ij}, g_j), & if\ RN(v_{ij}) < RN(g_j) \end{cases} \text{for cost type criteria (See Section 2.2)} \end{cases} \tag{30b}$$

$$d_E(v_{ij}, g_j) = \begin{cases} \sqrt{\left(v_{ij}^L - g_j^U\right)^2 + \left(v_{ij}^U - g_j^L\right)^2}, & for\ benefit\ type\ criteria \\ \sqrt{\left(v_{ij}^L - g_j^L\right)^2 + \left(v_{ij}^U - g_j^U\right)^2}, & for\ cost\ type\ criteria \end{cases} \tag{31}$$

where $\left[g_j^L, g_j^U\right]$ is the border approximation area for criterion $C_j$ ($j = 1, 2, \dots, n$).

Now, an alternative $A_i$ will belong to the border approximation area $(G)$ if $q_{ij} = 0$, upper approximation area $(G^+)$ if $q_{ij} > 0$, and lower approximation area $(G^-)$ if $q_{ij} < 0$. The ideal alternative $(A^+)$ can be found in the upper approximation area $(G^+)$ whereas the lower approximation area $(G^-)$ contains the anti-ideal alternative $(A^-)$ (See Fig. 2). For alternative $A_i$ to be best, it is necessary to have as many criteria belonging to the upper approximation area$(G^+)$.

$Step$ 6. Determine the ranking orders of all alternatives.

Alternative $A_i$ is near or equal to the ideal and anti-Ideal solution if the distance value $q_{ij} \in G^+ \ and \ G^-$ respectively. For criteria function values of the alternative sites, the distances of the alternatives from BAA vector are added. Summing row elements of matrix$(Q)$, final score values of the criterion functions for alternatives are obtained as (32):

$$S(A_i) = \sum_{j=1}^{n} q_{ij},\ i = 1, 2, \dots, m. \tag{32}$$

The alternatives are ranked as per $S(A_i)$. The flowchart of the proposed method is given in figure 3.

## 4. Case study: Ranking Indian cities based on medical tourism

The proposed hybrid MCDM methodology is applied for selecting the most appropriate cities in India for medical tourism. Primary data are collected using Interviews, questionnaires and observations of the admitted patients in the hospitals for the year 2014-15. Secondary data, both official and business, are collected for information regarding medical providers, city's demography and government policies, from different expertise like policy makers, tour and hospitality managers and from medical professionals. After preliminary screening, we choose seven different maximizing criteria $\{C_i: i = 1,2, \dots,7\}$ (Table 2) are categorized into three operational groups based on major thrust areas. As per experts' opinion, nine major cities in India are chosen which presently excels in medical tourism, (shown in Table 3). The evaluation is based on a 9- point linguistic scale (Zhu et al., 2015).

*4.1 Result based on Rough AHP*

*Step 1.* Collect individual judgments of six decision makers and using of the rough AHP method, construct six non-negative pairwise comparison matrices and consistency ratio of each judgment matrix is calculated. Clearly $CR^i < 0.10$ for all $B^i$ $(i = 1, 2, \dots, 6)$. So, all the pairwise comparison matrices are acceptable.

*Step 2.* Next, these matrices are integrated to generate a rough comparison matrix (Table 4) using equations (1)-(10). The individual comparison matrices for six different experts are as follows:

$$B^1 = \begin{pmatrix} 1 & 5 & 3 & 1 & 3 & 5 & 9 \\ 1/5 & 1 & 1/3 & 1/5 & 1/3 & 1/3 & 7 \\ 1/3 & 3 & 1 & 1/5 & 1/3 & 3 & 5 \\ 1 & 5 & 5 & 1 & 3 & 3 & 9 \\ 1/3 & 3 & 3 & 1/3 & 1 & 3 & 7 \\ 1/5 & 3 & 1/3 & 1/3 & 1/3 & 1 & 3 \\ 1/9 & 1/7 & 1/5 & 1/9 & 1/7 & 1/3 & 1 \end{pmatrix} \quad B^2 = \begin{pmatrix} 1 & 5 & 3 & 3 & 1 & 5 & 7 \\ 1/5 & 1 & 1/3 & 1/5 & 1/5 & 1/3 & 5 \\ 1/3 & 3 & 1 & 1/5 & 1/3 & 3 & 7 \\ 1/3 & 5 & 5 & 1 & 3 & 5 & 9 \\ 1 & 5 & 3 & 1/3 & 1 & 3 & 7 \\ 1/5 & 3 & 1/3 & 1/5 & 1/3 & 1 & 3 \\ 1/7 & 1/5 & 1/7 & 1/9 & 1/7 & 1/3 & 1 \end{pmatrix} \quad B^3 = \begin{pmatrix} 1 & 5 & 3 & 1 & 1 & 3 & 9 \\ 1/5 & 1 & 1 & 1/3 & 1/5 & 3 & 7 \\ 1/3 & 1 & 1 & 1/5 & 1/3 & 3 & 7 \\ 1 & 3 & 5 & 1 & 3 & 5 & 9 \\ 1 & 5 & 3 & 1/3 & 1 & 3 & 7 \\ 1/3 & 1/3 & 1/3 & 1/5 & 1/3 & 1 & 3 \\ 1/9 & 1/5 & 1/7 & 1/9 & 1/7 & 1/3 & 1 \end{pmatrix}$$

$CR^1 = 0.078 < 0.10$ $\quad$ $CR^2 = 0.090 < 0.10$ $\quad$ $CR^3 = 0.061 < 0.10$

$$B^4 = \begin{pmatrix} 1 & 3 & 3 & 1 & 5 & 5 & 7 \\ 1/3 & 1 & 1/3 & 1/5 & 1/5 & 1/3 & 3 \\ 1/3 & 3 & 1 & 1/3 & 3 & 5 & 5 \\ 1/ & 5 & 3 & 1 & 3 & 5 & 7 \\ 1/5 & 5 & 1/3 & 1/3 & 1 & 3 & 5 \\ 1/5 & 3 & 1/5 & 1/5 & 1/3 & 1 & 3 \\ 1/7 & 1/3 & 1/5 & 1/7 & 1/5 & 1/3 & 1 \end{pmatrix} \quad B^5 = \begin{pmatrix} 1 & 7 & 5 & 1 & 3 & 5 & 9 \\ 1/7 & 1 & 1/3 & 1/5 & 1/5 & 1/3 & 3 \\ 1/5 & 3 & 1 & 1/3 & 1/3 & 3 & 5 \\ 1 & 5 & 3 & 1 & 5 & 3 & 7 \\ 1/3 & 5 & 3 & 1/5 & 1 & 5 & 3 \\ 1/5 & 3 & 1/3 & 1/3 & 1/5 & 1 & 3 \\ 1/9 & 1/3 & 1/5 & 1/7 & 1/3 & 1/3 & 1 \end{pmatrix} \quad B^6 = \begin{pmatrix} 1 & 5 & 3 & 1 & 3 & 5 & 9 \\ 1/5 & 1 & 1/3 & 1/5 & 1/5 & 1/3 & 3 \\ 1/3 & 3 & 1 & 1/3 & 1/3 & 3 & 5 \\ 1 & 5 & 3 & 1 & 3 & 5 & 7 \\ 1/3 & 5 & 3 & 1/3 & 1 & 3 & 5 \\ 1/5 & 3 & 1/3 & 1/5 & 1/3 & 1 & 5 \\ 1/9 & 1/3 & 1/5 & 1/7 & 1/5 & 1/5 & 1 \end{pmatrix}$$

$CR^4 = 0.089 < 0.10$ $\quad$ $CR^5 = 0.0895 < 0.10$ $\quad$ $CR^6 = 0.0602 < 0.10$

Note: (1) 9-point scale system: 1 = very low; 3 = low; 5 = moderate; 7 = high; 9 = very high

(2) 2, 4, 6, 8 are intermediate values

(3) $\lceil \, , \rfloor$ represent rough number.

(4) Criteria set: $\{C_i : i = 1,2, \dots ,7\}$

*Step 3*. Computation of the criteria weights in rough number is done applying equations (20)-(21). Finally we normalize those weights according to the equations (22) to get normalized rough number valued weights (Table 5).

*4.2 Decision making using Rough MABAC*

Rough MABAC is adopted to determine the final ranking of Medical Tourism (MT) sites once we get the relative weights of the criteria set. Each expert gives a subjective and comprehensive judgment/evaluation for each alternative sites based on qualitative criteria under consideration. All the experts are supposed to use the same 9 point scale ranging from "very low to very high" for performance evaluation of a MT site, shown in Table 6.

*Step 1*. The original group decision data in Table 5 are translated into initial rough decision matrix $X = \left(\left[x_{ij}^{L}, x_{ij}^{U}\right]\right)_{9\times 7}$ (Table 7) using equations (1) – (12).

*Step 2*. Depending on the type of the criteria (cost type or minimizing and benefit type or maximizing) we first find the values of $x_j^{+}$ and $x_j^{-}$ according to equations (26) and (27). Next, all the entries of initial rough decision matrix (Table 6) are normalized using equations (24) and (25). Thus, the normalized rough group decision matrix $N = \left(\left[n_{ij}^{L}, n_{ij}^{U}\right]\right)_{9\times 7}$is computed (Table 8).

*Step 3*. Calculate the weighted rough group decision matrix (Table 9) by multiplying the corresponding normalized weights (table 4) with normalized rough group decision matrix (Table 8), applying eqn. (28).

*Step 4*. Using the geometric aggregation operation for interval valued numbers, the BAA matrix (Table 10) for each evaluation criterion is computed according to equations (29). For example

$$g_1^L = (1.246 * 1.387 * 1.072 * 1.072 * 0.882 * 0.882 * 0.788 * 1.211 * 0.788)^{\frac{1}{9}} = 1.017$$

$$g_1^U = (1.821 * 2.000 * 1.684 * 1.684 * 1.388 * 1.388 * 1.149 * 1.687 * 1.149)^{\frac{1}{9}} = 1.524$$

$$and\ g_1 = [1.017, 1.524]$$

*Step 5*.The distances of the alternative cities from BAA calculated to compute the distance matrix $Q$ (Table 11) according to the interval valued Euclidean distance operator [shown in equations (30) and (31)].

*Step 6*.The closeness coefficients/final score $S(A_i)$ of the alternatives sites to the ideal/optimal alternative site are calculated by equations (32). For example,

$$S(A_2) = 0.603 + 0.007 + 0.181 + 0.624 + 0.494 + 0.077 + 0.020 = 2.006$$

$$S(A_8) = 0.253 - 0.016 + 0.041 - 0.229 + 0.494 - 0.082 + 0.029 = 0.492$$

Ranking is done (Table 12) according to the better $S(A_i)$ value the better alternative. Here, $A_2$ turned out to be best choice.

*4.3 Comparisons with other two methods*

A comparative analysis of the proposed method is done with previously known rough TOPSIS (Song et al., 2014) and rough AHP-VIKOR (Zhu et al., 2015) for checking its validity. The results are computed in Table 13. Also to determine the validity of the obtained ranking for a given decision problem, the following criteria proposed by Wang and Triantaphyllou (2008), are also tested.

***Test criterion 1.*** *Keeping relative importance of each decision criteria same in an MCDM method, there will be no change in best position of alternative in replacing a non-optimal alternative by a worse alternative*.

The modified decision matrix is considered and relative weights of criteria are kept same. Here, we have interchanged the values of $A_3\ and\ A_1\ with\ A_5\ and\ \ A_6$ respectively, in the initial decision matrix and evaluated by the proposed method.

In this case, the obtained ranking is: $A_2 > A_5 > A_6 > A_8 > A_4 > A_9 > A_7 > A_1 > A_3$.

This result shows that ranking of alternatives by proposed rough MABAC method remains unchanged when a non-optimal alternative is replaced by another worse alternative.

***Test criterion 2***. *Decision making method should follow transitivity property*.

The original MCDM problem is de-composed into a set of smaller MCDM problems $\{A_1, A_2, A_3, A_4, A_5, A_6, A_7\}$ and$\{A_3, A_4, A_7, A_8, A_9\}$. Following the steps of proposed MABAC method, rankings $A_2 > A_3 > A_1 > A_4 > A_7 > A_6 > A_5$ and $A_3 > A_8 > A_4 > A_9 > A_7$ are obtained for smaller MCDM problems. Thus the transitive property for MCDM methods is verified for the proposed method here.

*Test criterion 3. Decomposing an original MCDM problem in ranking of alternatives, the combined ranking of the alternatives should be identical to the original ranking of undecomposed problem.*

If the rankings of the alternatives of sub-problems are combined together, the final ranking $A_2 > A_3 > A_1 > A_8 > A_4 > A_9 > A_7 > A_6 > A_5$ is similar to undecomposed MCDM problem.

## *5. Further Analysis*

From the above comparison it is clear that by using rough AHP-MABAC method, we obtain the same results with other two methods. The main reason of using MABAC method is the simple computation procedure and the stability (consistency) of solution (Pamucar et al. 2015; Xue et al. 2016). The MABAC method is a particularly pragmatic and reliable tool for rational decision making.

One more benefit of this method is that it enables us to visualize of performance and assessment of individual MT sites as per each criteria and vice versa (Table 16 and 17). It shows the pair-wise comparison between each alternative's performances and the ideal value of each criterion. From the distance matrix we can directly conclude whether an alternative performs better than the ideal value in the considered problem. But TOPSIS and VIKOR methods do not produce such a direct observation.

As per result shown in Table (16)-(17), we can directly apprehend the performances of the nine alternative sites with respect to seven criteria belonging to three operational groups: Infrastructure, Medical Tourism Services, planning and policies. Dimension '*Infrastructure*', has three criteria class, namely, Quality of infrastructure of Health Care Institution ($C_1$), Transportation Convenience and city demography ($C_2$) and Informational Infrastructure and distribution channels ($C_3$). Site $A_2$ belongs to upper approximation area according to all criteria whereas $A_1$and $A_3$ belong to upper approximation area according to all criteria except one. $A_1$ belongs to lower approximation area of $C_2$ while $A_3$ that of $C_6$. But, $A_3$ precedes $A_1$ in the final ranking table since $A_3$ gets cumulative advantage over $A_1$ according to all criteria. Similarly, evaluation can be done for others. A decision maker just looking at the Table (11) or (16) or (17) would be able to do the assessment of the alternative sites and suggest the MT authorities/stakeholders of Medical Tourism Sectors to take care of their planning, infrastructure, and services etc. for better performances and improvements.

## 6. Conclusion

In the present scenario, strong economic boost in infrastructure sector and availability of skilled personnel has paved a smooth way for development of medical tourism in India. As different cities in India acquiring global accreditation for medical tourism spots, analysis and selection of the most suitable one considering city demography and different interests of stakeholders, is the central problem of this paper. For this purpose, nine alternative cities and seven criteria are considered on the basis of experts' opinion. Criteria are clubbed up in three different operational groups (Infrastructure, medical services, and government policies) which prioritize the criteria, building an impact on evaluation and selection of the sites.

This study proposes an integrated rough AHP – MABAC method to facilitate a more precise analysis of the alternatives, considering several criteria in uncertain environment. Rough number is introduced to aggregate individual judgments, priorities and measure vagueness. Different relative weights of criteria is more realistic in many practical MCDM problems, especially in complex and uncertain environments. Rough AHP enables to measure consistency of preferences, manipulate multiple decision makers and calculate the relative importance for each criterion. On the other hand, MABAC possess simple computation procedure and the stability (consistency) of solution. Particularly, this method also divides the performances of alternatives into two groups: upper and lower approximation area of each criteria function. Here, we utilize rough MABAC to evaluate and classify the alternative cities into positive performer(s) and negative performer (s) in the distance matrix according to each and every criteria under consideration.

Let us consider the instance that the city Chennai ($A_2$) has been termed "*India's health capital*". Chennai attracts about 45 percent of health tourists from abroad arriving in the country and 30 to 40 percent of domestic health tourists (Hamid, 2012). Despite its super-specialty hospitals and world class health infrastructure, Bangalore ($A_1$) is far behind Chennai in attracting international patients, due to much better flight connectivity of Chennai to United States (USA), Middle East and other gulf countries. This is where Bangalore's medical tourism industry is lagging behind (Indian Express, May 19, 2013). From our analysis it evident that, Bangalore ($A_2$) scores negatively in Transportation Convenience ($C_2$) in the distance matrix (Table 11). So, Chennai needs to keep the present performance and Bangalore must focus to improve on Transportation Convenience ($C_2$) to attract more medical tourists. Similar arguments can be done for other sites.

Hence, the proposed method successfully is applied to rough numbers under incomplete data, effectively avoiding vague and ambiguous judgments. Hence, the proposed method is more practical, realistic, comprehensive and applicable for any multi-criteria group decision making in an uncertain environment. In future, rough MABAC would produce interesting hybrid MCDM methods with the combination of other MCDM techniques like, ANP, DEMATEL-ANP, and Shannon Entropy etc.

**References**

1. Brans, J. P., & Vincke, P. (1985). Note—A Preference Ranking Organisation Method: (The PROMETHEE Method for Multiple Criteria Decision-Making). *Management science*, 31(6), 647-656.
2. Chakraborty, J (2006). Medical tourism: a growth engine for foreign exchange earnings. *Health care management.* Available from URL http:/ www.healthcaremanagement.corn.
3. Charnes, A., Cooper, W. W., & Rhodes, E. (1978). Measuring the efficiency of decision making units. *European journal of operational research*, 2(6), 429-444.
4. Connell, J. (2006). Medical tourism: sea, sun, sand and Surgery. *Tourism Management*. 27, 1093-1100.
5. Connell, J. (2013). Contemporary Medical tourism: Conceptualization, culture and commodification, *Tourism Management*, 34, 1-13.
6. de Korvin, A., McKeegan, C., & Kleyle, R. (1998). Knowledge acquisition using rough sets when membership values are fuzzy sets. *Journal of Intelligent & Fuzzy Systems*, 6(2), 237-244.
7. Eshghi, K., & Nematian, J. (2008). Special classes of mathematical programming models with fuzzy random variables. *Journal of Intelligent & Fuzzy Systems*, 19(2), 131-140.
8. Greco, S., Matarazzo, B., &Slowinski, R. (2001). Rough set theory for multi-criteria decision analysis. *EuropeanJournal of Operational Research*, 129(1), 1–47.
9. Gordon, J., & Shortliffe, E. H. (1984). The Dempster-Shafer theory of evidence. Rule-Based Expert Systems: *The MYCIN Experiments of the Stanford Heuristic Programming Project*, 3, 832-838.

10. Hennig, C., Meila, M., Murtagh, F., &Rocci, R. (Eds.). (2015). Handbook of cluster analysis. CRC Press.
11. Hwang, C. L., & Yoon, K. (1981). Multiple attribute decision making: methods and applications, in: *Lecture Notes in Economic and Mathematical Systems,* Spring, Berlin, Germany.
12. Joshi, D., & Kumar, S. (2016). Interval-valued intuitionistic hesitant fuzzy Choquet integral based TOPSIS method for multi-criteria group decision making, *European Journal of Operational Research* 248, 183–191.
13. Julong, D. (1989). Introduction to grey system theory. *The Journal of grey system*, 1(1), 1-24.
14. Khoo L, Tor S, & Zhai L. (1999). A rough-set based approach for classification and rule induction. *International Journal of Advanced Manufacturing Technology*, 15(6), 438–444.
15. Opricovic, S. (1998). Multicriteria optimization of civil engineering systems. *Faculty of Pennsylvania*, Belgrade.
16. Pamučar, D., & Ćirović, G. (2015). The selection of transport and handling resources in logistics centers using Multi-Attributive Border Approximation area Comparison (MABAC). *Expert Systems with Applications*, 42(6), 3016-3028.
17. Pawlak Z. (1982). Rough sets. *International Journal of Computing and Information Sciences*, 11(5), 341–356.
18. Qin, J., Liu, X., & Pedrycz, W. (2015). An extended VIKOR method based on prospect theory for multiple attribute decision making under interval type-2 fuzzy environment. *Knowledge-Based Systems*, 86, 116-130.
19. Roy, B. (1968). Classement et Choix en Presence de Points de vue Multiples (la method Electre). Revue Francaise d'Informatique et de *Recherche Operationnelle* 8 (1): 57–75.
20. Saaty, T. L. (1977). A scaling method for priorities in hierarchical structures. *Journal of mathematical psychology*, 15(3), 234-281.
21. Saaty, T. L. (1990). How to make a decision: the analytic hierarchy process. *European journal of operational research*, 48(1), 9-26.
22. Saaty, T. L., & Vargas, L. G. (2012). Models, methods, concepts & applications of the analytic hierarchy process (Vol. 175). *Springer Science & Business Media*.
23. Saaty, T. L. (1996). Decision making with dependence and feedback: The analytic network process. *RWS Publication*.
24. Shafer, G. (1976). A mathematical theory of evidence (Vol. 1, pp. xiii+-297). *Princeton: Princeton university press.*
25. Song, W., Ming, X., & Wu, Z. (2013). An integrated rough number-based approach to design concept evaluation under subjective environments. *Journal of Engineering Design*, 24(5), 320-341.

26. Song, W., Ming, X., Wu, Z., & Zhu, B. (2014). A rough TOPSIS approach for failure mode and effects analysis in uncertain environments. *Quality and Reliability Engineering International*, 30(4), 473-486.
27. Wang, X., & Triantaphyllou, E. (2008). Ranking irregularities when evaluating alternatives by using some ELECTRE methods. *Omega*, 36, 45–63.
28. Xue, Y. X., You, J. X., Lai, X. D., & Liu, H. C. (2016). An interval-valued intuitionistic fuzzy MABAC approach for material selection with incomplete weight information. *Applied Soft Computing*, 38, 703-713.
29. Zadeh, L. (1965). Fuzzy sets. *Information and Control* 8 (3): 338–53.
30. Zhai, L. Y., Khoo, L. P., & Zhong, Z. W. (2008). A rough set enhanced fuzzy approach to quality function deployment. *The International Journal of Advanced Manufacturing Technology*, 37(5-6), 613-624.
31. Zhai, L. Y., Khoo, L. P., & Zhong, Z. W. (2009). A rough set based QFD approach to the management of imprecise design information in product development. *Advanced Engineering Informatics*, 23(2), 222-228.
32. Zhu, G. N., Hu, J., Qi, J., Gu, C. C., & Peng, Y. H. (2015). An integrated AHP and VIKOR for design concept evaluation based on rough number. *Advanced Engineering Informatics*, 29(3), 408-418.
33. Hamid, Zubeda (20 August 2012). "The medical capital's place in history". The Hindu (Chennai: The Hindu). Retrieved 15 Sep 2012.
34. Kalyanam, S. City scores low on medical tourism, Indian Express, 19th May 2013

**Table 1.** Random Consistency Index (RI) (Saaty and Vargas, 2012)

| n | 3 | 4 | 5 | 6 | 7 | 8 | 9 | 10 |
|---|---|---|---|---|---|---|---|---|
| RI | 0.52 | 0.89 | 1.11 | 1.25 | 1.35 | 1.40 | 1.45 | 1.49 |

**Table 2**. Framework for operational groups and Evaluation criteria in medical tourism

| Operational groups and Evaluating Criteria | Major thrust areas. |
|---|---|
| ***Strengthening of Infrastructure*** | |
| $C_1$. Quality of Infrastructure of HealthCare Institutions | Accreditation and Certification; Private-Pubic-Partnership participation |
| $C_2$. Transportation Convenience and City demography. | Socio-Economic-Political Condition; Regulatory Conditions and Attributions; Distance from Major Airport, railway stations; Economic Expenditure and activities. |
| $C_3$. Informational Infrastructure and distribution channels | Web-based health information; Social, print and e-media; Overseas and national campaigns through trade fairs. |
| ***Strengthening of Medical-Tourism Services*** | |
| $C_4$. Supply of skilled Human resources and new job creations | Special Expertise and training personnel; Reputation and Recommendation |
| $C_5$. Quality of Medical Operator and consultation Centers | Certification and Accreditations; Cultural and Ethical perspectives; Strong partnerships with major chain Hospitals and clinics for scheduled appointment; Linkage with Hotels, Airlines and providing assistance in getting medical visa. |
| ***Planning and policies towards Medical Tourism.*** | |
| $C_6$. Schedule planning and packaging of Medical Tourism. | Development of Major Hotels and market outlook; Promotion of Heath resort, Spa and Medi-clinics; Promotion of City tours and Eco-tourism; Identify the target population; Identify specific areas for medical tourism. |
| $C_7$. Government laws and policies towards Medical Tourism. | Health Care policies from Ministry of Health/ Hospital; Tourism Policies to promote Medical tourism; Incentives for Investment in Medical sector; Law enforcement and amendments to relevant laws; Policy and legal issues easy and systematic. |

**Table 3**: Brief description of the conceptual alternatives

| Alternatives | Name of Cities | Accreditation | | | | Tie-up of Medical centers with Tour operators | State-Govt. Policy and approach |
|---|---|---|---|---|---|---|---|
| | | Medical Centers | | Tour Operators | | | |
| | | JCI | NABH | MTQA | GOI | | |
| $A_1$ | Bangalore | 3 | 19 | 1 | 4 | High | Good |
| $A_2$ | Chennai | 2 | 9 | 0 | 2 | Very High | Very Good |
| $A_3$ | Delhi | 5 | 41 | 1 | 15 | High | Very Very Good |
| $A_4$ | Hyderabad | 1 | 23 | 0 | 1 | Low | Good |
| $A_5$ | Jaipur | 1 | 7 | 0 | 1 | Very low | Moderate |
| $A_6$ | Kolkata | 1 | 6 | 1 | 3 | Low | Moderate |
| $A_7$ | Mumbai | 3 | 12 | 0 | 4 | Moderate | Moderate |
| $A_8$ | Pune | 1 | 9 | 0 | 1 | High | Good |
| $A_9$ | Kochi | 0 | 3 | 1 | 1 | High | Moderate |

**Table 4:** Aggregated rough comparison matrix

| | $C_1$ | $C_2$ | $C_3$ | $C_4$ | $C_5$ | $C_6$ | $C_7$ |
|---|---|---|---|---|---|---|---|
| $C_1$ | [1.000, 1.000] | [4.400, 5.600] | [3.056, 3.611] | [1.056, 1.611] | [1.878, 3.472] | [4.389, 4.944] | [7.889, 8.778] |
| $C_2$ | [0.185, 0.242] | [1.000, 1.000] | [0.352, 0.537] | [0.204, 0.241] | [0.204, 0.241] | [0.407, 1.148] | [3.489, 5.222] |
| $C_3$ | [0.293, 0.330] | [2.389, 2.944] | [1.000, 1.000] | [0.233, 0.300] | [0.407, 1.148] | [3.056, 3.611] | [5.222, 6.111] |
| $C_4$ | [0.796, 0.982] | [4.389, 4.944] | [3.500, 4.500] | [1.000, 1.000] | [3.056, 3.611] | [3.889, 4.778] | [7.500, 8.500] |
| $C_5$ | [0.361,0.722] | [4.389, 4.944] | [2.185, 2.926] | [0.293, 0.330] | [1.000, 1.000] | [3.056, 3.611] | [4.778, 6.511] |
| $C_6$ | [0.204, 0.241] | [2.185, 2.926] | [0.293, 0.330] | [0.215, 0.274] | [0.293, 0.330] | [1.000, 1.000] | [3.056, 3.611] |
| $C_7$ | [0.115, 0.129] | [0.213, 0.303] | [0.168, 0.194] | [0.119, 0.135] | [0.159, 0.234] | [0.293, 0.330] | [1.000, 1.000] |

**Source-Indian Tourism Statistics, 2011, 2013; www.ibef.org; IMaCS Research; Bureau of Immigration, India.*

**Table 5:** Aggregated weights and normalized weights

| | $C_1$ | $C_2$ | $C_3$ | $C_4$ | $C_5$ | $C_6$ | $C_7$ |
|---|---|---|---|---|---|---|---|
| Rough weights: | [2.652, 3.367] | [0.452, 0.643] | [1.008,1.330] | [2.716, 3.168] | [1.469,1.873] | [0.590, 0.692] | [0.217, 0.259] |
| Normalized rough weights: | [0.788, 1.000] | [0.134, 0.191] | [0.300, 0.395] | [0.807, 0.941] | [0.436, 0.556] | [0.175,0.205] | [0.065, 0.077] |

**Table 6:** Experts' based decision matrix

| | Experts | $C_1$ | $C_2$ | $C_3$ | $C_4$ | $C_5$ | $C_6$ | $C_7$ | | Experts | $C_1$ | $C_2$ | $C_3$ | $C_4$ | $C_5$ | $C_6$ | $C_7$ |
|---|---|---|---|---|---|---|---|---|---|---|---|---|---|---|---|---|---|
| $A_1$ | 1 | 9 | 7 | 7 | 5 | 7 | 7 | 7 | $A_6$ | 1 | 5 | 5 | 5 | 7 | 7 | 7 | 3 |
| | 2 | 7 | 5 | 9 | 7 | 7 | 7 | 7 | | 2 | 7 | 5 | 5 | 7 | 5 | 7 | 3 |
| | 3 | 7 | 7 | 5 | 5 | 7 | 7 | 7 | | 3 | 7 | 7 | 5 | 5 | 7 | 5 | 7 |
| | 4 | 9 | 7 | 7 | 5 | 7 | 7 | 7 | | 4 | 5 | 5 | 5 | 7 | 7 | 7 | 3 |
| | 5 | 7 | 7 | 7 | 7 | 7 | 7 | 5 | | 5 | 7 | 5 | 5 | 5 | 5 | 5 | 7 |
| | 6 | 7 | 5 | 5 | 5 | 5 | 5 | 7 | | 6 | 5 | 5 | 7 | 5 | 5 | 5 | 5 |
| $A_2$ | 1 | 9 | 9 | 9 | 7 | 9 | 7 | 7 | $A_7$ | 1 | 5 | 5 | 5 | 5 | 7 | 7 | 3 |
| | 2 | 9 | 7 | 9 | 7 | 7 | 7 | 7 | | 2 | 5 | 7 | 7 | 7 | 7 | 7 | 3 |
| | 3 | 9 | 7 | 9 | 5 | 9 | 7 | 5 | | 3 | 5 | 5 | 7 | 7 | 5 | 7 | 7 |
| | 4 | 9 | 9 | 9 | 7 | 9 | 7 | 7 | | 4 | 5 | 5 | 5 | 5 | 7 | 7 | 3 |
| | 5 | 7 | 9 | 9 | 7 | 7 | 7 | 5 | | 5 | 7 | 7 | 7 | 7 | 7 | 7 | 5 |
| | 6 | 7 | 7 | 7 | 5 | 7 | 5 | 7 | | 6 | 5 | 5 | 7 | 7 | 5 | 5 | 5 |
| $A_3$ | 1 | 7 | 9 | 7 | 9 | 9 | 5 | 7 | $A_8$ | 1 | 7 | 5 | 9 | 5 | 7 | 3 | 5 |
| | 2 | 7 | 7 | 9 | 9 | 7 | 5 | 7 | | 2 | 7 | 7 | 7 | 5 | 7 | 3 | 5 |
| | 3 | 7 | 9 | 5 | 9 | 9 | 5 | 7 | | 3 | 7 | 5 | 9 | 7 | 5 | 7 | 5 |
| | 4 | 7 | 9 | 7 | 9 | 9 | 5 | 7 | | 4 | 7 | 5 | 9 | 5 | 7 | 3 | 5 |
| | 5 | 9 | 9 | 5 | 5 | 5 | 5 | 7 | | 5 | 9 | 7 | 9 | 7 | 7 | 7 | 5 |
| | 6 | 5 | 7 | 7 | 5 | 5 | 7 | 5 | | 6 | 7 | 5 | 9 | 7 | 7 | 5 | 7 |
| $A_4$ | 1 | 7 | 7 | 7 | 5 | 3 | 5 | 5 | $A_9$ | 1 | 5 | 7 | 5 | 7 | 5 | 5 | 5 |
| | 2 | 7 | 5 | 7 | 7 | 5 | 5 | 5 | | 2 | 5 | 7 | 7 | 7 | 5 | 5 | 5 |
| | 3 | 7 | 7 | 5 | 5 | 7 | 5 | 7 | | 3 | 5 | 7 | 7 | 5 | 7 | 7 | 7 |

| | | | | | | | | | | | | | | | | |
|---|---|---|---|---|---|---|---|---|---|---|---|---|---|---|---|---|
| | 4 | 7 | 7 | 7 | 5 | 3 | 5 | 5 | 4 | 5 | 7 | 5 | 7 | 5 | 5 | 5 |
| | 5 | 9 | 5 | 5 | 5 | 5 | 5 | 7 | 5 | 7 | 7 | 7 | 7 | 7 | 7 | 5 |
| | 6 | 5 | 5 | 7 | 7 | 5 | 7 | 7 | 6 | 5 | 7 | 7 | 5 | 7 | 7 | 7 |
| A5 | 1 | 5 | 5 | 3 | 7 | 5 | 3 | 3 | | | | | | | | |
| | 2 | 7 | 3 | 5 | 7 | 5 | 3 | 3 | | | | | | | | |
| | 3 | 7 | 7 | 7 | 5 | 7 | 7 | 7 | | | | | | | | |
| | 4 | 5 | 5 | 3 | 7 | 5 | 3 | 3 | | | | | | | | |
| | 5 | 7 | 5 | 5 | 5 | 5 | 5 | 7 | | | | | | | | |
| | 6 | 5 | 5 | 5 | 7 | 7 | 5 | 5 | | | | | | | | |

* 9-point scale system: 1 = very low; 3 = low; 5 = moderate; 7 = high; 9 = very high

**Table 7:** Initial rough decision matrix (X)

| | $C_1$ | $C_2$ | $C_3$ | $C_4$ | $C_5$ | $C_6$ | $C_7$ |
|---|---|---|---|---|---|---|---|
| $A_1$ | [7.222, 8.111] | [5.222, 6.111] | [6.389, 6.944] | [6.500, 6.944] | [5.878, 7.472] | [6.389, 6.944] | [6.389, 6.944] |
| $A_2$ | [7.889, 8.778] | [5.889, 6.778] | [7.500, 8.500] | [7.889, 8.778] | [8.389, 8.944] | [6.389, 6.944] | [5.889, 6.778] |
| $A_3$ | [6.400, 7.600] | [6.778, 8.556] | [6.278, 8.361] | [8.389, 8.944] | [5.878, 7.472] | [5.056, 5.611] | [6.389, 6.944] |
| $A_4$ | [6.400, 7.600] | [5.222, 6.111] | [3.878, 5.472] | [5.500, 6.500] | [5.889, 6.778] | [5.056, 5.611] | [5.500, 6.500] |
| $A_5$ | [5.500, 6.500] | [5.889, 6.778] | [5.222, 6.111] | [5.056, 5.611] | [3.878, 5.472] | [3.489, 5.222] | [3.639, 5.722] |
| $A_6$ | [5.500, 6.500] | [5.500, 6.500] | [5.500,6.500] | [5.056, 5.611] | [5.056, 5.611] | [5.500, 6.500] | [3.639, 5.722] |
| $A_7$ | [5.056, 5.611] | [5.889, 6.778] | [5.889, 6.778] | [5.222, 6.111] | [5.889, 6.778] | [6.389, 6.944] | [3.489, 5.222] |
| $A_8$ | [7.056, 7.611] | [5.500, 6.500] | [6.389, 6.944] | [5.222, 6.111] | [8.389, 8.944] | [3.639, 5.722] | [6.389, 6.944] |
| $A_9$ | [5.056, 5.611] | [5.889, 6.778] | [5.500, 6.500] | [7.000, 7.000] | [5.889, 6.778] | [5.500, 6.500] | [5.222, 6.111] |

**Table 8:** Normalized rough decision matrix (N)

| | $C_1$ | $C_2$ | $C_3$ | $C_4$ | $C_5$ | $C_6$ | $C_7$ |
|---|---|---|---|---|---|---|---|
| $A_1$ | [0.582 , 0.821] | [0.000, 0.267] | [0.543, 0.664] | [0.343, 0.486] | [0.395, 0.709] | [0.839, 1.000] | [0.839, 1.000] |
| $A_2$ | [0.761, 1.000] | [0.200, 0.467] | [0.784, 1.000] | [0.729, 0.957] | [0.890, 1.000] | [0.839, 1.000] | [0.695, 0.952] |
| $A_3$ | [0.361, 0.684] | [0.467, 1.000] | [0.519, 0.970] | [0.857, 1.000] | [0.395, 0.709] | [0.453, 0.614] | [0.839, 1.000] |
| $A_4$ | [0.361, 0.684] | [0.000, 0.267] | [0.000, 0.345] | [0.114,0.371] | [0.397, 0.572] | [0.453, 0.614] | [0.582, 0.871] |
| $A_5$ | [0.119, 0.388] | [0.200, 0.467] | [0.291, 0.483] | [0.000, 0.143] | [0.000, 0.315] | [0.000, 0.502] | [0.043, 0.646] |
| $A_6$ | [0.119, 0.388] | [0.083, 0.383] | [0.351, 0.567] | [0.000,0.143] | [0.233, 0.342] | [0.582, 0.871] | [0.043, 0.646] |
| $A_7$ | [0.000, 0.149] | [0.200, 0.467] | [0.435, 0.627] | [0.043, 0.271] | [0.397,0.572] | [0.839, 1.000] | [0.000, 0.502] |
| $A_8$ | [0.537,0.687] | [0.083, 0.383] | [0.543,0.664] | [0.043, 0.271] | [0.890, 1.000] | [0.043, 0.646] | [0.839, 1.000] |
| $A_9$ | [0.000, 0.149] | [0.200, 0.467] | [0.351, 0.567] | [0.500, 0.500] | [0.397, 0.572] | [0.582, 0.871] | [0.502, 0.759] |

**Table 9:** Weighted decision matrix (V)

| | $C_1$ | $C_2$ | $C_3$ | $C_4$ | $C_5$ | $C_6$ | $C_7$ |
|---|---|---|---|---|---|---|---|
| $A_1$ | [1.246, 1.821] | [0.134, 0.242] | [0.462, 0.657] | [1.083, 1.398] | [0.609, 0.951] | [0.323, 0.411] | [0.119, 0.154] |
| $A_2$ | [1.387, 2.000] | [0.161, 0.280] | [0.534, 0.790] | [1.394, 1.841] | [0.825, 1.113] | [0.323, 0.411] | [0.109, 0.150] |
| $A_3$ | [1.072, 1.684] | [0.197, 0.382] | [0.455, 0.778] | [1.498, 1.882] | [0.609, 0.951] | [0.255, 0.332] | [0.119, 0.154] |
| $A_4$ | [1.072, 1.684] | [0.134, 0.242] | [0.300, 0.531] | [0.899, 1.290] | [0.610, 0.875] | [0.255, 0.332] | [0.102, 0.144] |
| $A_5$ | [0.882, 1.388] | [0.161, 0.280] | [0.387, 0.586] | [0.807, 1.075] | [0.436, 0.731] | [0.175, 0.308] | [0.067, 0.127] |
| $A_6$ | [0.882, 1.388] | [0.145, 0.264] | [0.405,0.619] | [0.807, 1.075] | [0.538, 0.747] | [0.277, 0.384] | [0.067, 0.127] |
| $A_7$ | [0.788, 1.149] | [0.161, 0.280] | [0.430, 0.643] | [0.841, 1.196] | [0.610, 0.875] | [0.323, 0.411] | [0.065, 0.116] |
| $A_8$ | [1.211, 1.687] | [0.145, 0.264] | [0.462, 0.657] | [0.841, 1.196] | [0.825,1.113] | [0.183, 0.338] | [0.119, 0.154] |
| $A_9$ | [0.788, 1.149] | [0.161, 0.280] | [0.405, 0.619] | [1.210, 1.411] | [0.610, 0.875] | [0.277, 0.384] | [0.097, 0.135] |

**Table 10**: Border approximation area (BAA) matrix (*g*)

| | $C_1$ | $C_2$ | $C_3$ | $C_4$ | $C_5$ | $C_6$ | $C_7$ |
|---|---|---|---|---|---|---|---|
| $g_j$ | $[1.017, 1.524]$ | $[0.155, 0.277]$ | $[0.422, 0.649]$ | $[1.014, 1.347]$ | $[0.505, 0.736]$ | $[0.260, 0.366]$ | $[0.093, 0.139]$ |

**Table 11**: Distance of the alternative from the Border Approximation Area matrix.

| | $C_1$ | $C_2$ | $C_3$ | $C_4$ | $C_5$ | $C_6$ | $C_7$ |
|---|---|---|---|---|---|---|---|
| $A_1$ | 0.375 | -0.041 | 0.041 | 0.086 | 0.239 | 0.077 | 0.029 |
| $A_2$ | 0.603 | 0.007 | 0.181 | 0.624 | 0.494 | 0.077 | 0.020 |
| $A_3$ | 0.169 | 0.113 | 0.134 | 0.721 | 0.239 | -0.035 | 0.029 |
| $A_4$ | 0.169 | -0.041 | -0.170 | -0.128 | 0.174 | -0.035 | 0.010 |
| $A_5$ | -0.191 | 0.007 | -0.072 | -0.342 | -0.069 | -0.102 | -0.029 |
| $A_6$ | -0.191 | -0.016 | -0.034 | -0.342 | 0.035 | 0.026 | -0.029 |
| $A_7$ | -0.439 | 0.007 | 0.010 | -0.229 | 0.174 | 0.077 | -0.037 |
| $A_8$ | 0.253 | -0.016 | 0.041 | -0.229 | 0.494 | -0.082 | 0.029 |
| $A_9$ | -0.439 | 0.007 | -0.034 | 0.206 | 0.174 | 0.026 | -0.006 |

**Table 12**: Final score and ranking by rough MABAC

| Alternatives | Final Score $S(A_i)$ | Ranking |
|---|---|---|
| $A_1$ | 0.807 | 3 |
| $A_2$ | 2.006 | 1 |
| $A_3$ | 1.371 | 2 |
| $A_4$ | -0.020 | 5 |
| $A_5$ | -0.797 | 9 |
| $A_6$ | -0.552 | 8 |
| $A_7$ | -0.437 | 7 |
| $A_8$ | 0.492 | 4 |
| $A_9$ | -0.066 | 6 |

**Table 13**: Comparisons with other methods

| Methods | Ranking orders |
|---|---|
| Rough TOPSIS method (Song et al. 2014) | $A_2 > A_3 > A_1 > A_8 > A_4 > A_9 > A_7 > A_6 > A_5$ |
| Rough AHP-VIKOR method (Zhu et al. 2015) | $A_2 > A_3 > A_1 > A_8 > A_4 > A_9 > A_7 > A_6 > A_5$ (Based on $Q$ values). |
| Proposed Rough AHP-MABAC | $A_2 > A_3 > A_1 > A_8 > A_4 > A_9 > A_7 > A_6 > A_5$ |

**Table 14**: Assessment of Medical Tourism cities as per criteria's based on Rough AHP-MABAC model.

| Operational groups | Criteria | Position of Indian cities as per Criteria taken | |
|---|---|---|---|
| | | Upper Approximation Area | Lower Approximation Area |
| Infrastructure | $C_1$: Quality of Infrastructure of Healthcare Institutions | $A_1$: Bangalore<br>$A_2$: Chennai<br>$A_3$: Delhi<br>$A_4$: Hyderabad<br>$A_8$: Pune | $A_5$: Jaipur<br>$A_6$: Kolkata<br>$A_7$: Mumbai<br>$A_9$: Kochi |
| | $C_2$: Transportation Convenience and City Demography | $A_2$: Chennai<br>$A_3$: Delhi<br>$A_5$: Jaipur<br>$A_7$: Mumbai<br>$A_9$: Kochi | $A_1$: Bangalore<br>$A_4$: Hyderabad<br>$A_6$: Kolkata<br>$A_8$: Pune |
| | $C_3$: Informational Infrastructure and Distribution channels. | $A_1$: Bangalore<br>$A_2$: Chennai<br>$A_3$: Delhi<br>$A_7$: Mumbai<br>$A_8$: Pune | $A_4$: Hyderabad<br>$A_5$: Jaipur<br>$A_6$: Kolkata<br>$A_9$: Kochi |
| Medical Tourism Services | $C_4$: Supply of skilled human resources and new job creations | $A_1$: Bangalore<br>$A_2$: Chennai<br>$A_3$: Delhi<br>$A_9$: Kochi | $A_4$: Hyderabad<br>$A_5$: Jaipur<br>$A_6$: Kolkata<br>$A_7$: Mumbai<br>$A_8$: Pune |
| | $C_5$: Quality of Medical Operator and Consultation Centers. | $A_1$: Bangalore<br>$A_2$: Chennai<br>$A_3$: Delhi<br>$A_4$: Hyderabad<br>$A_6$: Kolkata<br>$A_7$: Mumbai<br>$A_9$: Kochi<br>$A_8$: Pune | $A_5$: Jaipur |
| Planning and Policies | $C_6$: Schedule planning and packaging of Medical Tourism | $A_1$: Bangalore<br>$A_2$: Chennai<br>$A_6$: Kolkata<br>$A_7$: Mumbai<br>$A_9$: Kochi | $A_3$: Delhi<br>$A_4$: Hyderabad<br>$A_5$: Jaipur<br>$A_8$: Pune |
| | $C_7$: Regional Government laws and policies towards Medical Tourism | $A_1$: Bangalore<br>$A_2$: Chennai<br>$A_3$: Delhi<br>$A_4$: Hyderabad<br>$A_8$: Pune | $A_5$: Jaipur<br>$A_6$: Kolkata<br>$A_7$: Mumbai<br>$A_9$: Kochi |

**Table 15**: Assessment of Indian cities categorized in operational groups based on benefit and risk criteria.

| Indian cities | Operational groups | Upper Approximation Area | Lower Approximation Area |
|---|---|---|---|
| | Infrastructure | $C_1$, $C_3$ | $C_2$ |
| $A_1$: Bangalore | Medical-Tourism Services | $C_4$, $C_5$ | |
| | Planning and policies | $C_6$, $C_7$ | |
| | Infrastructure | $C_1$, $C_2$, $C_3$ | |
| $A_2$: Chennai | Medical-Tourism Services | $C_4$, $C_5$ | |
| | Planning and policies | $C_6$, $C_7$ | |
| | Infrastructure | $C_1$, $C_2$, $C_3$ | |
| $A_3$: Delhi | Medical-Tourism Services | $C_4$, $C_5$ | |
| | Planning and policies | $C_7$ | $C_6$ |
| | Infrastructure | $C_1$ | $C_2$, $C_3$ |
| $A_4$: Hyderabad | Medical-Tourism Services | $C_5$ | $C_4$ |
| | Planning and policies | $C_7$ | $C_6$ |
| | Infrastructure | $C_2$ | $C_1$, $C_3$ |
| $A_5$: Jaipur | Medical-Tourism Services | | $C_4$, $C_5$ |
| | Planning and policies | | $C_6$, $C_7$ |
| | Infrastructure | | $C_1$, $C_2$, $C_3$ |
| $A_6$: Kolkata | Medical-Tourism Services | $C_5$ | $C_4$ |
| | Planning and policies | $C_6$ | $C_7$ |
| | Infrastructure | $C_2$, $C_3$ | $C_1$ |
| $A_7$: Mumbai | Medical-Tourism Services | $C_5$ | $C_4$ |
| | Planning and policies | $C_6$ | $C_7$ |
| | Infrastructure | $C_1$, $C_3$ | $C_2$ |
| $A_8$: Pune | Medical-Tourism Services | $C_5$ | $C_4$ |
| | Planning and policies | $C_7$ | $C_6$ |
| | Infrastructure | $C_2$ | $C_1$, $C_3$ |
| $A_9$: Kochi | Medical-Tourism Services | $C_4$, $C_5$ | |
| | Planning and policies | $C_6$ | $C_7$ |

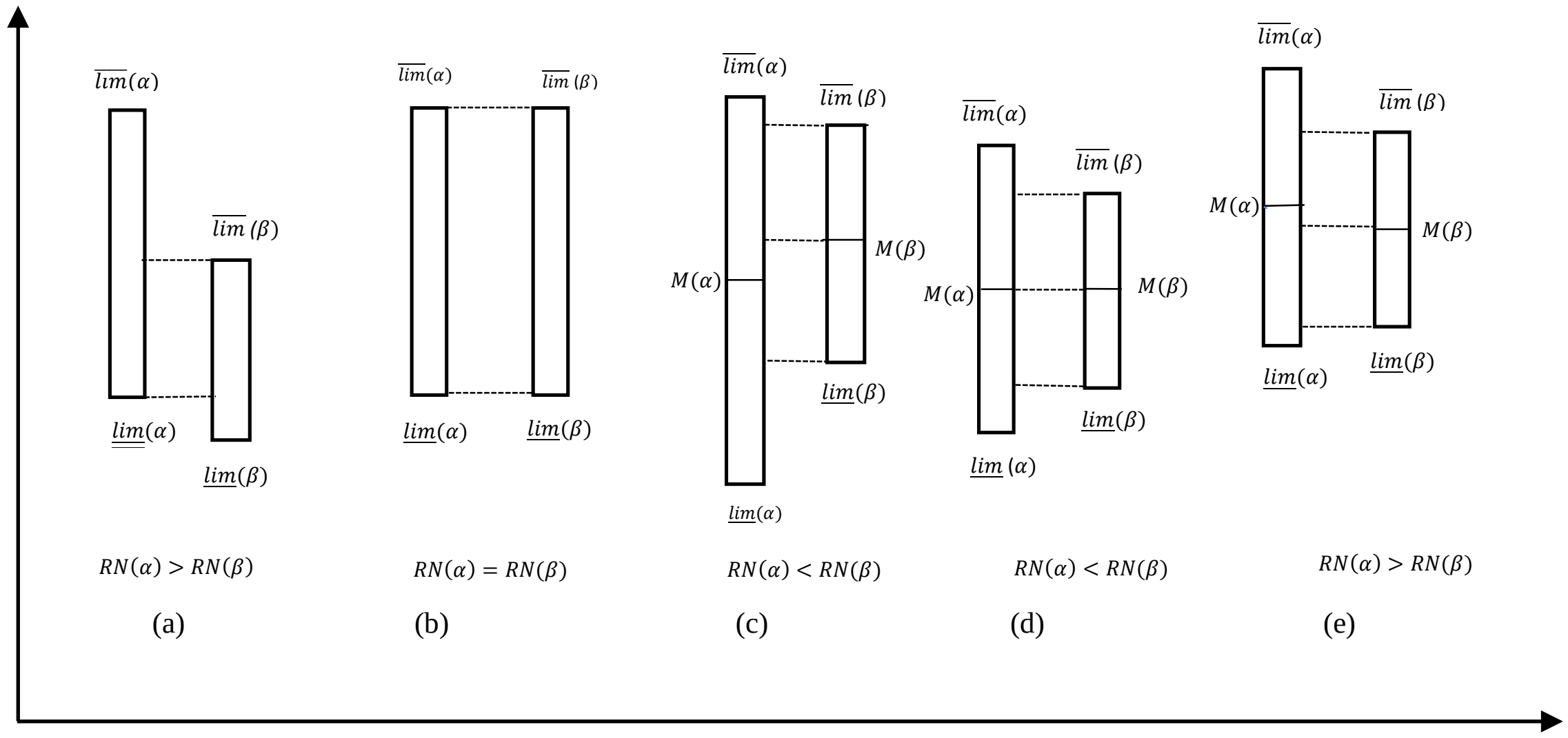

**Fig. 1:** Comparison of Rough Numbers (Zhai et al. 2008)

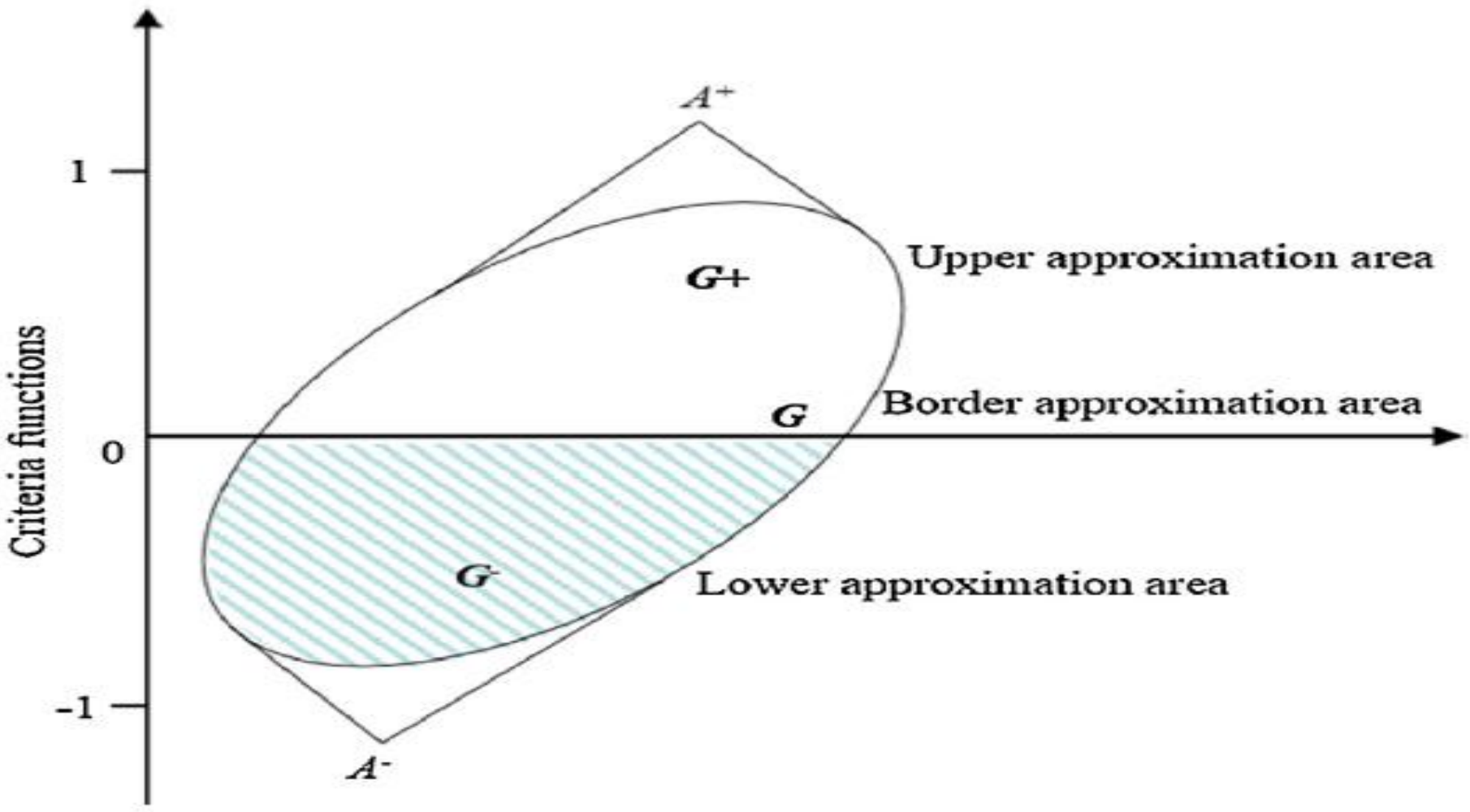

**Fig. 2:** Arrangement of $G^+$, $G^-$, and $G$ approximation area (Xue et al., 2016).

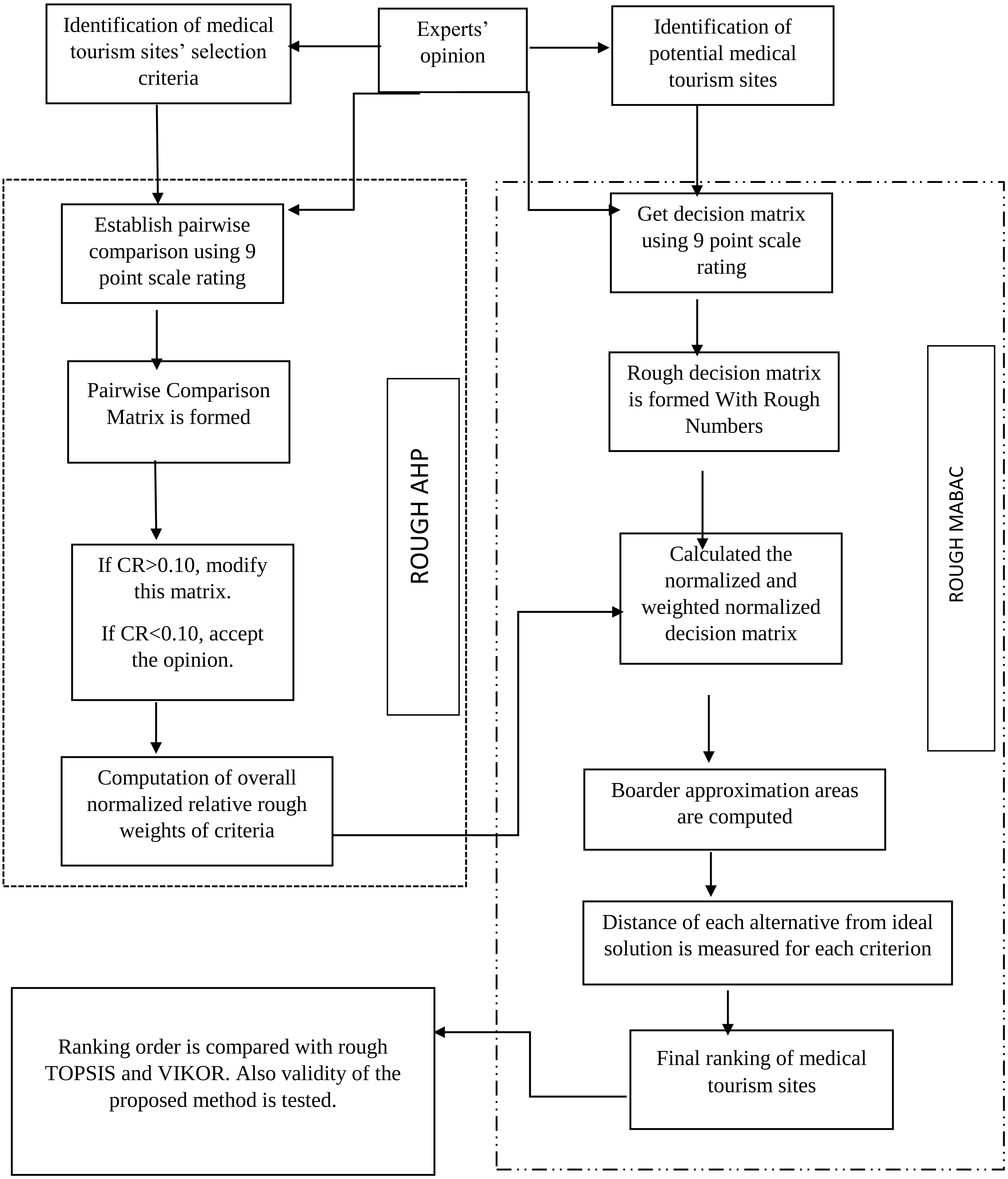

**Fig 3.** Flowchart of the proposed method

## Appendix 1: Feedback form for priority of criteria

| | Quality of Infrastructure of HealthCare Institutions | Transportation Convenience and City demography | Informational Infrastructure and distribution channels | Supply of skilled Human resources and new job | Quality of Medical Operator and consultation | Schedule planning and packaging of Medical | Government laws and policies towards |
|---|---|---|---|---|---|---|---|
| Quality of Infrastructure of HealthCare Institutions | Equal priority | High | Moderate | Equal priority | High | Low | Very high |
| Transportation Convenience and City demography | | Equal priority | | | | | |
| Informational Infrastructure and distribution channels | | | Equal priority | | | | |
| Supply of skilled Human resources and new job creations | | | | Equal priority | | | |
| Quality of Medical Operator and consultation Centers | | | | | Equal priority | | |
| Schedule planning and packaging of Medical Tourism. | | | | | | Equal priority | |
| Government laws and policies towards Medical Tourism. | | | | | | | Equal priority |

## Appendix 2: Feedback form for performance rating of potential medical tourism sites based on criteria

| | Quality of Infrastructure of HealthCare Institutions | Transportation Convenience and City demography | Informational Infrastructure and distribution channels. | Supply of skilled Human resources and new job | Quality of Medical Operator and consultation | Schedule planning and packaging of Medical | Government laws and policies towards Medical Tourism. |
|---|---|---|---|---|---|---|---|
| $A_1$ : Bangalore | High | Very High | | | | | |
| $A_2$ : Chennai | | | | | | | |
| $A_3$ : Delhi | | | | | | | |
| $A_4$ : Hyderabad | | | Moderate | | | | |
| $A_5$ : Jaipur | | | | Low | Good | | |
| $A_6$ : Kolkata | | | | | | High | |
| $A_7$ :Mumbai | | | | | | | Good |
| $A_8$ : Pune | | | | | | | |
| $A_9$ : Kochi | | | | | | | |